\documentclass[sigconf]{acmart}

\AtBeginDocument{%
  }

\copyrightyear{2022}
\acmYear{2022}
\setcopyright{acmcopyright}\acmConference[CIKM '22]{Proceedings of the 31st ACM International Conference on Information and Knowledge Management}{October 17--21, 2022}{Atlanta, GA, USA}
\acmBooktitle{Proceedings of the 31st ACM International Conference on Information and Knowledge Management (CIKM '22), October 17--21, 2022, Atlanta, GA, USA} \acmPrice{15.00}
\acmDOI{10.1145/3511808.3557232}
\acmISBN{978-1-4503-9236-5/22/10}

\newcommand{\model}{{{Fed\_BVA}}}
\usepackage{multirow,color}
\usepackage{cleveref}
\usepackage[font=small]{caption}
\usepackage{booktabs}
\usepackage{subfigure}
\usepackage{caption,enumitem}
\usepackage{tabularx}
\usepackage{algorithm}
\usepackage{algorithmic}
\makeatletter
\newcommand\footnoteref[1]{\protected@xdef\@thefnmark{\ref{#1}}\@footnotemark}
\makeatother

\allowdisplaybreaks

\begin{document}
\setlength{\belowdisplayskip}{0.pt} \setlength{\belowdisplayshortskip}{0.pt}
\setlength{\abovedisplayskip}{1pt} \setlength{\abovedisplayshortskip}{1pt}

% \title{Adversarial Robustness through Bias Variance Decomposition: A New Perspective for Federated Learning}
\title[Adversarial Robustness through Bias Variance Decomposition: A New Perspective for Federated Learning]{Adversarial Robustness through Bias Variance Decomposition: \\ A New Perspective for Federated Learning}

\author{Yao Zhou}
\authornote{Both authors contributed equally to this research.}
\affiliation{%
  \institution{Instacart \& University of Illinois at Urbana Champaign}
  \city{}
  \country{}}
\email{yaozhou3@illinois.edu}

\author{Jun Wu}
\authornotemark[1]
\affiliation{%
  \institution{University of Illinois at Urbana Champaign}
  \city{}
  \country{}}
\email{junwu3@illinois.edu}

\author{Haixun Wang}
\affiliation{%
  \institution{Instacart}
  \city{}
  \country{}}
\email{haixun@gmail.com}

\author{Jingrui He}
\affiliation{%
  \institution{University of Illinois at Urbana Champaign}
  \city{}
  \country{}}
\email{jingrui@illinois.edu}

% \renewcommand{\shortauthors}{Zhou et al.}
\renewcommand{\shortauthors}{Yao Zhou, Jun Wu, Haixun Wang, \& Jingrui He}

\begin{abstract}
    Federated learning learns a neural network model by aggregating the knowledge from a group of distributed clients under the privacy-preserving constraint. In this work, we show that this paradigm might inherit the adversarial vulnerability of the centralized neural network, i.e., it has deteriorated performance on adversarial examples when the model is deployed. This is even more alarming when federated learning paradigm is designed to approximate the updating behavior of a centralized neural network. To solve this problem, we propose an adversarially robust federated learning framework, named {\tt \model}, with improved server and client update mechanisms. This is motivated by our observation that the generalization error in federated learning can be naturally decomposed into the bias and variance triggered by multiple clients' predictions. Thus, we propose to generate the adversarial examples via maximizing the bias and variance during server update, and learn the adversarially robust model updates with those examples during client update. As a result, an adversarially robust neural network can be aggregated from these improved local clients' model updates. The experiments are conducted on multiple benchmark data sets using several prevalent neural network models, and the empirical results show that our framework is robust against white-box and black-box adversarial corruptions under both IID and non-IID settings.
\end{abstract}







\maketitle

\vspace{-3mm}
\section{Introduction}
The explosive amount of decentralized user data collected from the ever-growing usage of smart devices, e.g., smartphones, wearable devices, home sensors, etc., has led to a surge of interest in the field of decentralized learning. To protect the privacy-sensitive data of local clients, federated learning~\citep{mcmahan2017communication, yang2019federated} has been proposed. It only allows a group of clients to train local models using their own data on each individual device, and then collectively merges the model updates on a central server using secure aggregation~\citep{homomorphic_encryption}. Due to its high privacy-preserving property, federated learning has attracted much attention in recent years along with the prevalence of efficient light-weight deep models~\citep{mobilenet} and low-cost network communications~\citep{konevcny2016federated}.

Most existing federated learning approaches~\cite{karimireddy2020scaffold,robust_agg_fed,Wang2020Federated} focus on improving the strategies of local model training (e.g., local SGD~\cite{stich2018local}) and global aggregation (e.g., FedAvg~\citep{mcmahan2017communication}). The intuition is to approximate the updating behavior of the centralized model trained on all clients' data. However, little effort has been devoted to systematically analyzing the adversarial robustness of federated learning paradigm. This becomes even more alarming when centralized machine learning models have been shown to be vulnerable to adversarial attacks when those models are deployed in the testing phase~\cite{goodfellow2014explaining,moosavi2016deepfool}. It is studied~\cite{zizzo2020fat,reisizadeh2020robust,shah2021adversarial} that a heuristic solution is to leverage the adversarial training techniques~\cite{madry2018towards,defense_quanquangu} for clients' local training. However, it suffers from expensive computation for local clients with limited storage and computational resources.

Our work studies the adversarial robustness of federated learning paradigm by investigating the generalization error incurred in the server's aggregation process from the perspective of bias-variance decomposition~\citep{pedro2000unified, valentini2004bias}. Specifically, we show that this generalization error on the central server can be decomposed as the combination of \textbf{bias} (triggered by the main prediction of these clients) and \textbf{variance} (triggered by the variations among clients' predictions). This motivates us to propose a novel adversarially robust federated learning framework {\tt \model}\footnote{\url{https://github.com/jwu4sml/FedBVA}}. The key idea is to perform the local robust training on clients by supplying them with bias-variance perturbed examples generated from a tiny auxiliary training set on the central server. It has the following advantages. First, it encourages the clients to consistently produce the optimal prediction for perturbed examples, thereby leading to a better generalization performance. Second, local adversarial training on the perturbed examples learns a robust local model, and thus an adversarially robust global model could be aggregated from clients' local updates. The experiments are conducted on neural networks with cross-entropy loss, however, other loss functions are also applicable as long as their gradients w.r.t. bias and variance are tractable to be computed. 

It is worth noting that our problem setting fundamentally differs from the existing Byzantine-robust federated learning~\citep{data2021byzantine,li2021ditto,robust_agg_fed,blanchard2017machine}. To be more specific, those works proposed to improve the robustness of federated learning against Byzantine failures induced by corrupted clients' updates during model training, by performing client-level robust training or designing server-level aggregation variants with hyper-parameter tuning. In contrast, we focus on the adversarial robustness of federated learning against adversarial examples, when the model is deployed in the testing phase. Generally, the problem studied in this paper assumes that the learning process is clean (no malicious behaviors or Byzantine faults are observed in clients). But we also empirically show in Subsection~\ref{sec:byzantine_attacks} that when this assumption is violated, our {\tt \model} can be robust against both adversarial perturbation and Byzantine failures by incorporating Byzantine-robust aggregation variants~\cite{blanchard2017machine,yin2018byzantine,chen2017distributed}.

Compared with previous work, our major contributions include:
\begin{itemize}[leftmargin=*,noitemsep,nolistsep]
    \item We provide the exact solution of bias-variance analysis w.r.t. the generalization error for neural networks in the federated learning setting. As a comparison, performing the adversarial training on conventional federated learning methods can only focus on the bias of the central model but ignore the variance.
    \item We demonstrate that the conventional federated learning framework is vulnerable to strong attacks with increasing communication rounds even if the adversarial training using the locally generated adversarial examples is performed on each client.
    \item Without violating the clients' privacy, we show that providing a tiny amount of bias-variance perturbed data from the central server to the clients through asymmetrical communication could dramatically improve the robustness of the training model under various adversarial settings.
\end{itemize}

The rest of this paper is organized as follows. The related work is reviewed in Section~\ref{sec:related_work}. In Section~\ref{sec:Preliminaries}, the preliminary is provided. Then, in Section~\ref{sec:framework}, a generic framework for adversarially robust federated learning is proposed, followed by the instantiated algorithm \textit{\tt \model} with bias-variance attacks in Section~\ref{sec:Algorithm}. The experiments are presented in Section~\ref{sec:experiments}. We conclude this paper in Section~\ref{sec:conclusion}.

\vspace{-3mm}
\section{Related Work}\label{sec:related_work}
% In this section, we introduce the related work on federated learning and bias-variance decomposition.

% \vspace{-3mm}
\subsection{Federated Learning}
Federated learning~\cite{konevcny2016federated,mcmahan2017communication,FedTriNet_fenglong_bigdata21,multi-task-fedml-fenglong-bigdata19} aggregates the knowledge from a large number of local devices (e.g., smartphone~\cite{hard2018federated}) with limited storage and computational resources. This aggregated knowledge can thus be leveraged to train the modern machine learning models (e.g., deep neural networks)~\cite{karimireddy2020scaffold,Wang2020Federated, dongqi_kdd_22_evolvegraph}. However, federated learning algorithms over neural networks are approximating the update dynamics of centralized neural networks, thus resulting in inheriting the adversarial vulnerability of centralized neural networks~\cite{goodfellow2014explaining,szegedy2014intriguing,wu2021indirect,zhou2019misc,zheng2021deep,DBLP:conf/kdd/ZhouZ0H20,yao_pure_kdd21,dongqi_cikm20_multiviewadversarial}. More specifically, the deployed neural network models in federated learning might incorrectly classify the adversarial examples generated by the existing evasion attacks (e.g., FGSM~\cite{goodfellow2014explaining}, PGD~\cite{madry2018towards}). 

There are two lines of related works for improving the robustness of federated learning. One is to develop robust model aggregation schemes~\citep{li2021ditto,data2021byzantine,robust_agg_fed,blanchard2017machine,yin2018byzantine,chen2017distributed} against Byzantine failures induced by corrupted client's updates during model training. Those works focus on performing client-level robust training or designing server-level aggregation variants with hyper-parameter tuning. The other one is to improve the model robustness against the adversarial attacks after model deployment~\cite{zizzo2020fat,shah2021adversarial}. This idea fundamentally differs from the Byzantine-robust approaches in the following. (1) Different from Byzantine-robust model training, it is assumed that no malicious behaviors or Byzantine faults are observed in clients in training. (2) It improves the model robustness by refining the decision boundary around the adversarial examples, whereas Byzantine-robust models focus on mitigating the impact of potential clients' faults. In this paper, we will mainly focus on the second scenario by analyzing the adversarial robustness of federated learning from the perspective of bias-variance analysis. Compared to previous works~\cite{zizzo2020fat,shah2021adversarial}, the bias-variance analysis on local clients allows capturing the global model dynamics for improving the adversarial robustness of the aggregated global model in the federated learning setting.

\vspace{-2mm}
\subsection{Bias-Variance Decomposition}
Bias-variance decomposition~\citep{geman1992neural,kohavi1996bias} was originally introduced to analyze the generalization error of a learning algorithm. Then, a generalized bias-variance decomposition \citep{pedro2000unified,valentini2004bias} was studied in the classification setting which enabled flexible loss functions (e.g., squared loss, zero-one loss). Specifically, it is shown that under mild conditions, the generalization error of a learning algorithm can be decomposed into bias, variance and irreducible noise. In this case, bias measures how the trained models over different training data sets deviate from the optimal model, and variance measures the differences among those trained models. Moreover, it is previously observed~\cite{geman1992neural} that the bias decreases and the variance increases monotonically with respect to the model complexity. It thus suggests that a better trade-off of bias and variance could lead to improving the generalization performance of a learning algorithm. Nevertheless, in recent years, it is empirically shown~\cite{vgg_zisserman_2015} that increasing the model complexity of deep neural networks tends to produce better generalization performance, which is in contradiction to previous bias-variance analysis. Following this idea, bias-variance trade-off was experimentally revisited on modern neural network models~\citep{belkin2019reconciling,yang2020rethinking}. It is found that for deep neural networks, the variance is more likely to increase first and then decrease with respect to the model complexity.
We would like to point out that compared to standard supervised learning, federated learning can be better characterized by the bias-variance decomposition. That is because the trained models in local clients can be naturally applied to define the bias and variance in the generalization error of a learning algorithm. To the best of our knowledge, this is the first work studying the federated learning problem from the perspective of bias-variance analysis.

\vspace{-2mm}
\section{Preliminaries}\label{sec:Preliminaries}
In this section, we formally present the problem definition and the bias-variance trade-off in the classification setting.

\vspace{-2mm}
\subsection{Federated learning}\label{subsec:fl}
In federated learning~\cite{mcmahan2017communication, yang2019federated}, there is a central server and $K$ different clients, each with access to a private training set $\mathcal{D}_k= \{(x_i^k, t_i^k)\}_{i=1}^{n_k}$. Here, $x_i^k$, $t_i^k$, and $n_k$ are the features, label, and number of training examples in the $k^{\mathrm{th}}$ client where $k=1,\cdots,K$. Each data $\mathcal{D}_k$ is exclusively owned by client $k$ and will not be shared with the central server or other clients. 

The goal of standard federated learning is to learn the prediction function by aggregating the knowledge of user data from a set of local clients without leaking user privacy. A typical framework~\citep{mcmahan2017communication,yang2019federated} of federated learning can be summarized as follows: (1) {\em Client Update:} Each client updates local parameters $w_k$ by minimizing the empirical loss over its own training set; (2) {\em Forward Communication:} Each client uploads its parameter update to the central server; (3) {\em Server Update:} It synchronously aggregates the received parameters; (4) {\em Backward Communication:} The global parameters are sent back to the clients.

\vspace{-2mm}
\subsection{Problem Definition and Motivation}\label{sec:problem_definition}
It can be seen that the neural network in federated learning actually approximates the updating behavior of a centralized model. For example, the update behavior of FedAvg~\citep{mcmahan2017communication} is closely equivalent to the stochastic gradient descent (SGD) on the centralized data when each client only takes one step of gradient descent. That explains why it might potentially inherit the adversarial vulnerability of centralized neural network models~\cite{goodfellow2014explaining,szegedy2014intriguing}. Therefore, in this paper, we study the adversarial robustness of neural networks in the federated learning setting. Formally, given a set of private training data $\{\mathcal{D}_k\}_{k=1}^K$ on $K$ different clients, a learning algorithm $f(\cdot)$ and loss function $L(\cdot, \cdot)$, {\bf\em adversarially robust federated learning} aims to output a trained model on the central server that is robust against adversarial perturbations on the test set $\mathcal{D}_{test}$.

It is previously~\cite{zizzo2020fat,reisizadeh2020robust,shah2021adversarial} revealed that one can simply apply the adversarial training techniques~\cite{madry2018towards} to clients' local training. Nevertheless, it is unclear how the local adversarial training on clients affects the adversarial robustness of the aggregated global model on the central server. To answer this question, we study the intrinsic relationship between the set of local clients' models and the aggregated server's model from the perspective of bias-variance trade-off. It is found that the generalization error of the server's model is induced by both bias and variance of local clients' models. Therefore, instead of separately focusing on the adversarial robustness of individual clients~\cite{zizzo2020fat,shah2021adversarial}, we propose to analyze the adversarially robust federated learning in a unified framework. The crucial idea is to generate some global adversarial examples shared with clients, by leveraging a tiny auxiliary training set $\mathcal{D}_s=\{(x_j^s, t_j^s)\}_{j=1}^{n_s}$ with $n_s$ ($n_s \ll \sum_{k=1}^K n_k$) examples on the central server. This will not break the privacy constraints, as the local data in every client is not shared. In real scenarios, the auxiliary data can be some representative or synthetic template examples. For example, during the COVID-19 pandemic, hospitals (local clients) would like to consider a few publicly accessible template data with typical symptoms\footnote{\scriptsize \url{https://www.cdc.gov/coronavirus/2019-ncov/symptoms-testing/symptoms.html}} for model training of the diagnostic system. Notice that federated semi-supervised learning~\cite{jeong2021federated} also considered a similar problem setting with labeled examples on the server and unlabeled examples on the clients. Instead, in our paper, we propose to improve the adversarial robustness of federated learned systems over local clients by taking those publicly accessible data $\mathcal{D}_s$ into consideration.

We would like to point out that our problem definition has the following properties. {\em (1) Asymmetrical communication:} The asymmetrical communication between each client and server cloud is allowed: the server provides both global model parameters and limited shared data to the clients; while each client uploads its local model parameters back to the server. This implies that compared to standard federated learning, the communication cost might increase due to those shared data, but it enables the improvement of adversarial robustness in federated learning (see Subsection~\ref{sec:shared_samples} for more empirical analysis)\footnote{\scriptsize We would like to leave the trade-off of communication cost and adversarial robustness as our future work. For example, asymmetrical communication can be device-dependent in real scenarios, by acquiring the transmitted data based on devices' storage and computational capacities.}. {\em (2) Data distribution:} All training examples on the clients and the server are assumed to follow the same data distribution. However, the experiments show that our proposed algorithm also achieves outstanding performance under the non-IID setting (see Subsection~\ref{sec:main_results}), which could be commonly seen among personalized clients in real scenarios. {\em (3) Shared learning algorithm:} All the clients are assumed to use the identical model $f(\cdot)$, including architectures as well as hyper-parameters (e.g., learning rate, local epochs, local batch size, etc.).

\vspace{-2mm}
\subsection{Bias-Variance Trade-off}

In this paper, we investigate the adversarially robust federated learning by studying the generalization error incurred in its aggregation process. We discover that the key to analyzing this error is from the perspective of the bias-variance trade-off. Following~\citep{pedro2000unified}, we define the optimal prediction, main prediction as well as the bias, variance, and noise for any real-valued loss function $L(\cdot,\cdot)$ as follows:

\begin{definition}\label{def:optimal_main}({\bf Optimal Prediction and Main Prediction}~\cite{pedro2000unified})
Given a loss function $L(\cdot, \cdot)$ and a learning algorithm $f(\cdot)$, optimal prediction $y_*$ and main prediction $y_m$ for input $x$ are defined as:
\begin{align*}
    y_*(x) = \arg\min_{y} \mathbb{E}_t[L(y,t)] \quad
    y_m(x) = \arg\min_{y'} \mathbb{E}_{\mathcal{D}}[L(f_{\mathcal{D}}(x), y')]
\end{align*}
where the label\footnote{\scriptsize In general, $t$ is a non-deterministic function~\citep{pedro2000unified, valentini2004bias} of $x$ when the irreducible noise is considered. Namely, if $x$ is sampled repeatedly, different values of $t$ will be observed.} $t$ and data set $\mathcal{D}$ are viewed as the random variables to denote the class label and training set, and $f_{\mathcal{D}}$ denotes the model trained on $\mathcal{D}$.
\end{definition}

\begin{definition}({\bf Bias, Variance and Noise})
Given a loss function $L(\cdot, \cdot)$ and a learning algorithm $f(\cdot)$, the bias, variance and noise can be defined as follows.
\begin{align*}
    B(x) = L(y_m, y_*), ~~
    V(x) = \mathbb{E}_{\mathcal{D}}[L(f_{\mathcal{D}}(x), y_m)],  ~~
    N(x) = \mathbb{E}_{t}[L(y_*, t)]
\end{align*}
Furthermore, there exists $\lambda, \lambda_0 \in \mathbb{R}$ such that the expected prediction loss $\mathbb{E}_{\mathcal{D},t}[L(f_{\mathcal{D}}(x), t)]$ for an example $x$ can be decomposed into bias, variance and noise as follows:
\begin{equation}\label{eq:bias_variance_noise}
\begin{aligned}
    \mathbb{E}_{\mathcal{D},t}[L(f_{\mathcal{D}}(x), t)] = B(x) + \lambda V(x) + \lambda_0 N(x)
\end{aligned}
\end{equation}

\end{definition}
In short, bias is the loss incurred by the main prediction w.r.t. the optimal prediction, and variance is the average loss incurred by all individual predictions w.r.t. the main prediction. Noise is conventionally assumed to be irreducible and independent of $f(\cdot)$. Our definitions of optimal prediction, main prediction, bias, variance and noise slightly differ from previous ones~\citep{pedro2000unified,valentini2004bias}. For example, conventional optimal prediction was defined as $y_*(x) = \arg\min_{y} \mathbb{E}_t[L(t, y)]$, and it is equivalent to our definition when loss function is symmetric over its arguments, i.e., $L(y_1, y_2) = L(y_2, y_1)$. 

\vspace{-2mm}
\section{The Proposed Framework}\label{sec:framework}
In this section, we present the adversarially robust federated learning framework. It follows the same paradigm of traditional federated learning (see Subsection~\ref{subsec:fl}) but with substantial modifications on the server update and client update as follows.

\vspace{-2mm}
\subsection{Server Update with Data Poisoning}
The server has two crucial components. The first one is model aggregation, which synchronously compresses and aggregates the received local model parameters. Another component is designed to produce adversarially perturbed examples which are induced by a poisoning attack algorithm for the usage of adversarial training.

\vspace{-1mm}
\subsubsection{Model Aggregation}
The overall goal of federated learning is to learn a prediction function by using knowledge from all clients such that it could generalize well over the test data set. When the central server receives the parameter updates from local clients, it aggregates the locally updated parameters to obtain a shared global model. It is notable that most existing federated learning approaches~\cite{mcmahan2017communication,Wang2020Federated,mohri2019agnostic} focus on developing advanced model aggregation schemes. One of the popular aggregation methods is FedAvg~\citep{mcmahan2017communication}, which aims to average the element-wise parameters of local models, i.e., $w_G = \mathrm{Aggregate}(w_1,\cdots,w_K) = \sum_{k=1}^K \frac{n_k}{n}w_k$ where $w_k$ is the model parameters in the $k^{\mathrm{th}}$ client and $n = \sum_{k=1}^K n_k$. In this paper, we focus on improving the adversarial robustness of federated learning by encouraging the local model parameters to be updated with adversarial examples (see next subsection). Then the adversarial robustness of local models can be naturally propagated into the server's global model after model aggregation. Therefore, our framework is flexible to be incorporated with existing model aggregation methods, e.g., FedAvg~\citep{mcmahan2017communication}, FedMA~\cite{Wang2020Federated}, AFL~\cite{mohri2019agnostic}, etc.

\vspace{-1mm}
\subsubsection{Adversarial Examples}
It is shown~\cite{goodfellow2014explaining} that the adversarial robustness of deep neural networks can be improved by updating the model parameters over adversarial examples. That is because the generalization error of neural networks on adversarial examples can be minimized during model training. However, it will encounter a few issues if we apply adversarial training on federated learning directly. Specifically, one intuitive solution for generating adversarial examples is to separately maximize the generalization error in each local client (see Subsection~\ref{sec:discussion} for more discussion). Its drawbacks are two folds: First, it will significantly increase the computational burden and memory usage on local clients. Second, the locally generated adversarial examples make the augmented data distributions of local clients much more biased~\cite{Xie2020Intriguing}, which challenges the standard server-level aggregation mechanisms~\cite{mohri2019agnostic}.

Instead, we study the adversarial robustness of federated learning from the perspective of bias-variance decomposition. 
The following theorem shows that in the classification setting, the generalization error of a learning algorithm can be decomposed into bias, variance, and irreducible noise. Note that this decomposition holds for any real-valued loss function in the binary setting~\citep{pedro2000unified} with a bias \& variance trade-off coefficient that has a closed-form expression.
\begin{theorem} \label{theorem:bvd_pedro}
In binary case, the decomposition in Eq.~(\ref{eq:bias_variance_noise}) is valid for any real-valued loss function that satisfy $\forall_y L(y,y) = 0$ and $\forall_{y_1 \neq y_2} L(y_1, y_2) \neq 0$ with $\lambda = 1$ if $y_m = y_*$ or $\lambda=-\frac{L(y_m,y_*)}{L(y_*, y_m)}$ otherwise. 
\end{theorem}

The proof of Theorem~\ref{theorem:bvd_pedro} is similar to~\citep{pedro2000unified}, we omit it here for space. Note that noise is irreducible and we usually drop this term in real applications~\citep{yang2020rethinking,pedro2000unified,valentini2004bias,geman1992neural}. Commonly used loss functions for this decomposition are square loss, zero-one loss, and cross-entropy loss with one-hot labels. For the multi-class setting, the closed-form solution of coefficient $\lambda$ for cross-entropy loss is given as follows.
\begin{equation}
    \lambda =
    \begin{cases}
      1 & \text{, \; $y_m = y_*$}\\
      -\frac{L(y_m,y_*)}{L(y_*, y_m)} & \text{, \; $y = y_m \;\text{or}\; y = y*$ (given $y_m \neq y_*$)}\\
      0 & \text{, \; $y \neq y_m \neq y_*$}
    \end{cases}    
\end{equation}
{\bf Remark.}
Intuitively, when $y_m = y_*$, no bias is induced and the generalization error mainly comes from variance; when bias exists $y_m \neq y_*$, a negative variance will help to reduce error when the prediction is identical to the main prediction or optimal prediction; otherwise, the variance is set to zero. In this paper, neural networks that use cross-entropy loss and mean squared loss with softmax prediction outputs are studied. Thus, we inherit their definition of bias \& variance directly, but treat the trade-off coefficient $\lambda$ as a hyper-parameter to tune because no closed-form expression of $\lambda$ is available in a general multi-class learning scenario.

Following~\citep{pedro2000unified,valentini2004bias}, we assume a noise-free scenario where the class label $t$ is a deterministic function of $x$ (i.e., if $x$ is sampled repeatedly, the same values of its class $t$ will be observed). The bias and variance can be empirically estimated over the clients' models. This motivates us to generate the adversarial examples by attacking the bias and variance induced by clients' models as:
\begin{equation} \label{eq:maximization}
    \max_{\hat{x}\in\Omega(x)} B(\hat{x}; w_1,\cdots,w_K) + \lambda V(\hat{x}; w_1,\cdots,w_K)
    \quad \forall (x,t)\in \mathcal{D}_s
\end{equation}
where $B(\hat{x}; w_1,\cdots,w_K)$ and $V(\hat{x}; w_1,\cdots,w_K)$ could be empirically estimated from a finite number of clients' parameters $w_1,\cdots,w_K$ trained on local training sets $\{\mathcal{D}_1, \cdots, \mathcal{D}_K\}$. Here $\lambda$ is a hyper-parameter to measure the trade-off of bias and variance, and $\Omega(x)$ is the perturbation constraint. $\hat{x}$ is the perturbed examples w.r.t. the clean example $x$ associated with class label $t$. Specifically, the perturbation constraint $\hat{x}\in \Omega(x)$ forces the adversarial example $\hat{x}$ to be visually indistinguishable w.r.t. $x$. In our paper, we use $l_{\infty}$-bounded adversaries~\citep{goodfellow2014explaining}, i.e., $\Omega(x) := \{\hat{x} \big| ||\hat{x} - x||_{\infty} \leq \epsilon \}$ for a perturbation magnitude $\epsilon$. 

Note that $\mathcal{D}_s$ (on the server) is the candidate subset of all available training examples that would lead to their perturbed counterparts. This is a more feasible setting as compared to generating adversarial examples on clients' devices because the server usually has powerful computational capacity in real scenarios that allows the usage of flexible poisoning attack algorithms. In this case, both poisoned examples and server model parameters would be sent back to each client ({\em Backward Communication}), while only clients' local parameters would be uploaded to the server ({\em Forward Communication}), i.e., the {\em asymmetrical communication}.

\vspace{-2mm}
\subsection{Robust Client Update}
The robust training of one client's prediction model (i.e., $w_k$) can be formulated as the following minimization problem.
\begin{equation} \label{eq:minimization}
    \min_{w_k} \left( \sum_{i=1}^{n_k} L(f_{\mathcal{D}_k}(x_i^k; w_k), t_i^k) + \sum_{j=1}^{n_s} L(f_{\mathcal{D}_k}(\hat{x}_j^s; w_k), t_j^s) \right)
\end{equation}
where $\hat{x}_j^s\in\Omega(x_j^s)$ is the perturbed example that is asymmetrically transmitted from the server. 

Intuitively, the bias measures the systematic loss of a learning algorithm, and the variance measures the prediction consistency of the learner over different training sets. Therefore, our robust federated learning framework has the following advantages: (i) it encourages the clients to consistently produce the optimal prediction for perturbed examples, thereby leading to a better generalization performance; (ii) local adversarial training on perturbed examples allows to learn a robust local model, and thus a robust global model could be aggregated from clients.

\vspace{-2mm}
\subsection{Discussion}\label{sec:discussion}
Theoretically, we could have another alternative robust federated learning strategy where the perturbed training examples of each client $k$ is generated on local devices from $\mathcal{D}_k$ instead of transmitted from the server. It is similar to~\citep{madry2018towards,tramer2018ensemble} where it iteratively synthesizes the adversarial counterparts of clean examples and updates the model parameters over perturbed training examples. Thus, each local robust model is trained individually. Nevertheless, poisoning attacks on the device will largely increase the computational cost and memory usage. Meanwhile, it only considers the client-specific loss and is still vulnerable against adversarial examples with increasing communication rounds. Both phenomena are observed in our experiments (see Fig.~\ref{fig_performance_fashion} and Fig.~\ref{fig_efficiency_fashion}).

\vspace{-2mm}
\section{Instantiated Algorithm}\label{sec:Algorithm}
In this section, we present the instantiated algorithm \textit{\tt \model}.% with bias-variance attacks.

\vspace{-2mm}
\subsection{Bias-Variance Attack} \label{BVD_gradients}
We first consider the maximization problem in Eq.~(\ref{eq:maximization}) using bias-variance based adversarial attacks. It finds the adversarial example $\hat{x}$ (from the original example $x$) that produces large bias and variance values w.r.t. clients' local models. To this end, we propose two gradient-based algorithms to generate adversarial examples.

{\bf Bias-variance based Fast Gradient Sign Method (BV-FGSM):} Following FGSM~\citep{goodfellow2014explaining}, it linearizes the maximization problem in Eq. (\ref{eq:maximization}) with one-step attack as follows.
\begin{equation}\label{BVD-FGSM}
\small
    \hat{x}_{\mathrm{BV-FGSM}} = x + \epsilon \cdot \mathrm{sign}\left(\nabla_x \left(B(x; w_1,\cdots,w_K) + \lambda V(x; w_1,\cdots,w_K) \right) \right)
\end{equation}
where $\mathrm{sign}(\cdot)$ denotes the sign function.

{\bf Bias-variance based Projected Gradient Descent (BV-PGD):} PGD~\citep{madry2018towards} can be considered as a multi-step variant of FGSM and might generate powerful adversarial examples. This motivated us to derive a BV-based PGD attack:
\begin{equation}\label{BVD-PGD}
\small
\begin{aligned}
    &\hat{x}_{\mathrm{BV-PGD}}^{l+1} \\ =& \mathrm{Proj}_{\Omega(x)} \left( \hat{x}^l + \epsilon \,\mathrm{sign}\left( \nabla_{\hat{x}^l} \left(B(\hat{x}^l; w_1,\cdots,w_K) + \lambda V(\hat{x}^l; w_1,\cdots,w_K) \right) \right) \right)
\end{aligned}
\end{equation}
where $\hat{x}^l$ is the adversarial example at the $l^{\mathrm{th}}$ step with the initialization $\hat{x}^0 = x$ and $\mathrm{Proj}_{\Omega(x)}(\cdot)$ projects each step onto $\Omega(x)$.

The proposed framework could b generalized to any gradient-based adversarial attack algorithms where the gradients of bias $B(\cdot)$ and variance $V(\cdot)$ w.r.t. $x$ are tractable when estimated from finite training sets. Compared with the existing attack methods~\citep{carlini2017towards,goodfellow2014explaining}, our loss function the adversary aims to optimize is a linear combination of bias and variance, whereas existing work mainly focused on attacking the overall classification error involving bias only.

\setlength{\textfloatsep}{4pt}
\begin{algorithm}[!t]
    % \centering
    \caption{\model}\label{algorithm:RobustFed}
    \begin{algorithmic}[1]
    \STATE {\bfseries Input:} $K$ (number of clients, with local data sets $\{\mathcal{D}_k\}_{k=1}^K$); $f$ (learning model); $L$ (loss function); $E$ (number of local epochs); $F$ (fraction of clients selected on each round); $B$ (batch size of local client); $\eta$ (learning rate); $\mathcal{D}_{s}$ (shared data set on server); $\epsilon$ (perturbation magnitude).
    \STATE {\bfseries Initialization:} Initialize $w_G^0$ and $\hat{\mathcal{D}}_s = {\mathcal{D}}_s$
    \FOR{each round $r=1,2,\cdots$}
    \STATE $m=\max(F\cdot K, 1)$
    \STATE $S_r \leftarrow$ randomly sampled $m$ clients
    \FOR{each client $k\in S_r$ in parallel}
        \STATE Initialize $k^{\mathrm{th}}$ client's model with $w_G^{r-1}$
        \STATE $\mathcal{B} \leftarrow$ split $\mathcal{D}_k \cup \hat{\mathcal{D}}_s$ into batches of size $B$
        \FOR{each local epoch $i=1,2,\cdots,E$}
        \FOR{local batch $(x, t) \in \mathcal{B}$}
        \STATE $w^{r}_k \leftarrow w^{r}_k - \eta \nabla_w L\left(f_{\mathcal{D}_k}(x; w^{r}_k), t\right)$
        \ENDFOR
        \ENDFOR
        \STATE Calculate $f_{\mathcal{D}_k}(x; w^{r}_k)$, $\nabla_x f_{\mathcal{D}_k}(x; w^{r}_k)$ for $\forall x \in \mathcal{D}_s$
    \ENDFOR
    \FOR{$(x, t) \in \mathcal{D}_s$}
    \STATE Estimate the gradients $\nabla_x B(x)$ and $\nabla_x V(x)$
    \STATE Update $\hat{x}\in \hat{\mathcal{D}}_s$ using Eq. (\ref{BVD-FGSM}) or (\ref{BVD-PGD}) 
    \ENDFOR
    \STATE $w_G^r \leftarrow$ {\bfseries Aggregate}($w_{k}^{r} | k\in S_r$)
    \ENDFOR
    \RETURN $w_G$
    \end{algorithmic}
\end{algorithm}

The following theorem states that bias $B(\cdot)$ and variance $V(\cdot)$ as well as their gradients over input $x$ could be estimated using the clients' models.
\begin{theorem}\label{thm:empirical_estimation}
Assume that $L(\cdot,\cdot)$ is the cross-entropy loss function, then, the empirical estimated main prediction $y_m$ for an input example $(x,t)$ has the following closed-form expression: 
$y_m(x) = \frac{1}{K}\sum_{k=1}^K f_{\mathcal{D}_k}(x; w_k)$.
Furthermore, the empirical bias and variance over an input $x$ are estimated as $B_{CE}(x) = \frac{1}{K} \sum_{k=1}^K  L(f_{\mathcal{D}_k}(x;w_k), t)$ and $V_{CE}(x) = H(y_m)$
and their corresponding gradients over $x$ are:
\begin{align*}
    \nabla_x B_{CE}(x) &= \frac{1}{K} \sum_{k=1}^K \nabla_x L(f_{\mathcal{D}_k}(x;w_k), t) \\
    \nabla_x V_{CE}(x) &= - \frac{1}{K}\sum_{k=1}^K \sum_{j=1}^C (\log{y_m^{(j)}} + 1)\cdot \nabla_{x} f_{\mathcal{D}_k}^{(j)}(x;w_k)
\end{align*}
\end{theorem}
\begin{proof}
Following the definition of main prediction, it aims to optimize $\arg\min_{y'} -\frac{1}{n}\sum_{k=1}^n \sum_{j=1}^C \big( f_{\mathcal{D}_k}^{(j)}(x) \cdot \log{y'^{(j)}} \big)$
with constraints $\sum_{j=1}^C y'^{(j)} = 1$ and $\sum_{j=1}^C f_{\mathcal{D}_k}^{(j)}(x) = 1$ for all $k=1,\cdots,n$. It can then be solved by the Lagrange multiplier.
\end{proof}

In the above theorem, $H(y_m) = -\sum_{j=1}^C y_m^{(j)}\log{y_m^{(j)}}$ is the entropy of the main prediction $y_m$ and $C$ is the number of classes. In addition, we also consider the case where $L(\cdot, \cdot)$ is the MSE loss function. Its main prediction $y_m$ for an input example $(x,t)$ has a closed-form expression which is exactly the same as the CE loss, its empirical bias and unbiased variance can only be estimated in the following formulas: 
\begin{equation*}
    \begin{aligned}
    B_{MSE}(x) &= \left|\left|\frac{1}{K} \sum_{k=1}^K f_{\mathcal{D}_k}(x; w_k) - t\right|\right|_2^2 \\
    V_{MSE}(x) &= \frac{1}{K-1} \sum_{k=1}^K  \left|\left|f_{\mathcal{D}_k}(x;w_k) - \frac{1}{K} \sum_{k=1}^K f_{\mathcal{D}_k}(x; w_k)\right|\right|_2^2
    \end{aligned}
\end{equation*}
and their gradients over input $x$ are:
\begin{align*}
    \nabla_x B_{MSE}(x) &= \left(\frac{1}{K} \sum_{k=1}^K f_{\mathcal{D}_k}(x; w_k) - t \right) \cdot  \left( \frac{1}{K} \sum_{k=1}^K \nabla_x f_{\mathcal{D}_k}(x; w_k) \right) \\
    \nabla_x V_{MSE}(x) &= \frac{1}{K-1} \sum_{k=1}^K \left( f_{\mathcal{D}_k}(x;w_k) - \frac{1}{K} \sum_{k=1}^K f_{\mathcal{D}_k}(x;w_k) \right) \cdot \\ 
    &\left(\nabla_x f_{\mathcal{D}_k}(x;w_k) - \frac{1}{K} \sum_{k=1}^K \nabla_x f_{\mathcal{D}_k}(x; w_k)\right)
\end{align*}

We observe that the empirical estimate of $\nabla_x B_{MSE}(x)$ ha higher computational complexity than $\nabla_x B_{CE}(x)$ because it involves the gradient calculation of prediction vector $f_{\mathcal{D}_k}(x; w_k)$ over the input tensor $x$. Besides, it is easy to show that the empirical estimate of $\nabla_x V_{MSE}(x)$ is also computationally expensive. A comparison between using CE and MSE losses is presented in Subsection~\ref{sec:mse_ce}.

\vspace{-2mm}
\subsection{ \model}
We present a novel robust \underline{fed}erated learning algorithm with our proposed \underline{b}ias-\underline{v}ariance \underline{a}ttacks, named \textit{\tt \model}. Following the framework defined in Eq.~(\ref{eq:maximization}) and Eq.~(\ref{eq:minimization}), key components of our algorithm are (1) bias-variance attacks for generating adversarial examples on the server, and (2) adversarial training using poisoned server examples together with clean local examples on each client. Therefore, we optimize these two objectives by producing the adversarial examples $\hat{\mathcal{D}}_s$ and updating the local model parameters $w$ iteratively. 

The proposed algorithm is summarized in Alg.~\ref{algorithm:RobustFed}. Given the server's $\mathcal{D}_{s}$ and clients' training data $\{\mathcal{D}_k\}_{k=1}^K$ as input, the output is a robust global model on the server. In this case, the clean server data $\mathcal{D}_{s}$ will be shared with all the clients. First, it initializes the server's model parameter $w_G$ and perturbed data $\hat{\mathcal{D}}_s$, and then assigns to the randomly selected clients (Steps 4-5). Next, each client optimizes its own local model (Steps 6-15) with the received global parameters $w_G$ as well as its own clean data $\mathcal{D}_k$, and uploads the updated parameters as well as the gradients of the local model on each shared server example back to the server. At last, the server generates the perturbed data $\hat{\mathcal{D}}_s$ (Step 16-19) using the proposed bias-variance attacks with aggregations (model parameters average, bias gradients average, and variance gradients average) in a similar manner as FedAvg~\citep{mcmahan2017communication}. These aggregations can be privacy secured if additive homomorphic encryption~\citep{homomorphic_encryption} is applied.

\vspace{-2mm}
\section{Experiments}\label{sec:experiments}
% In this section, we present the experimental results on evaluating the adversarial robustness of our proposed algorithm. 
% Code is available in this anonymous link\footnote{\tiny \url{https://drive.google.com/file/d/1oTrMv5rNSTZw0FN4vtHMONua-BBR7P1C/view?usp=sharing}}.

\vspace{-2mm}
\subsection{Settings}
\subsubsection{Data Sets}
We evaluate our proposed algorithm on four data sets: MNIST\footnote{\tiny \url{http://yann.lecun.com/exdb/mnist}}, Fashion-MNIST\footnote{\tiny \url{https://github.com/zalandoresearch/fashion-mnist}}, CIFAR-10\footnoteref{note1} and CIFAR-100\footnote{\tiny \label{note1} \url{https://www.cs.toronto.edu/~kriz/cifar.html}}. Following~\cite{mcmahan2017communication}, we consider two methods to partition the data over clients: IID and non-IID. For IID setting, the data is shuffled and uniformly partitioned into each client. For non-IID setting, data is divided into $2F\cdot K$ shards based on sorted labels, then assigns each client with 2 shards. Thereby, each client will have data with at most two classes.

\vspace{-1mm}
\subsubsection{Baselines}
The baseline models include: (1). {\tt Centralized}: the training with one centralized model, which is identical to the federated learning case that only has one client ($K=1$) with a fraction ($F=1$). (2). {\tt FedAvg}: the classical federated averaging model~\citep{mcmahan2017communication}. (3). {\tt FedAvg\_AT}: The simplified version of our proposed method where the local clients perform adversarial training with the asymmetrical transmitted perturbed data generated on top of {\tt FedAvg}'s aggregation. (4) - (6). {\tt Fed\_Bias}, {\tt Fed\_Variance}, {\tt Fed\_BVA}: Our proposed methods where the asymmetrical transmitted perturbed data is generated using the gradients of bias-only attack, variance-only attack, and bias-variance attack, respectively. (7). {\tt EAT}\footnote{\scriptsize Note that EAT shares the same idea as~\cite{zizzo2020fat} for robust federated learning.}: Ensemble adversarial training~\citep{tramer2018ensemble}, where each client performs local adversarial training, and their model updates are aggregated on the server using {\tt FedAvg}. (8). {\tt EAT+Fed\_BVA}: A combination of baselines (6) and (7). Note that the baselines (7) and (8) have high computational requirements on client devices, and are usually not preferred in real scenarios. For fair comparisons, all baselines are modified to the asymmetrical communication setting ({\tt FedAvg} and {\tt EAT} have clean $\mathcal{D}_s$ received), and all their initializations are set to be the same.

\begin{table}[!t]
\centering
\small
\setlength\tabcolsep{1.7pt}
\begin{tabular}{|l|c|c|c|c|}
\toprule
                  & MNIST & Fashion-MNIST & CIFAR-10 & CIFAR-100 (coarse) \\\midrule
\#   comms.       & 100   & 100           & 100      & 100                \\
\# clients (K)    & 100   & 100           & 20       & 20                 \\
fraction (F)      & 0.1   & 0.1           & 0.2      & 0.2                \\
\# epochs (E)      & 50    & 50            & 5        & 5                  \\
local batch   (B) & 64    & 64            & 128      & 128                \\
\# shared ($n_s$) & 64    & 64            & 30       & 60                 \\
\# categories     & 10    & 10            & 10       & 20         \\
\bottomrule
\end{tabular}
\caption{Learning setting of { \model}}
\label{setting:federated_learning}
\vspace{-5mm}
\end{table}

\vspace{-1mm}
\subsubsection{Model Configuration}
For {\tt \model} framework, we use a 4-layer CNN model for MNIST and Fashion-MNIST, and VGG9 architecture for CIFAR-10 and CIFAR-100. The training is performed using the SGD optimizer with a fixed learning rate of $0.01$ and momentum of value $0.9$. The trade-off coefficient between bias and variance is set to $\lambda=0.01$ for all experiments. All hyper-parameters of federated learning are presented in Table~\ref{setting:federated_learning}. We empirically demonstrate that these hyper-parameter settings are preferable in terms of both training accuracy and robustness (see the details of Fig.~\ref{fig:numShare} - Fig.~\ref{fig:localepoch} in Subsection~\ref{sec:Parameter_Analysis}). To demonstrate the robustness of our {\tt \model} framework, we evaluate the deployed server model on the test set $\mathcal{D}_{test}$ against adversarial attacks FGSM~\citep{goodfellow2014explaining}, PGD~\citep{madry2018towards} with 10 and 20 steps (i.e., PGD-10, PGD-20). Following~\citep{tramer2018ensemble, defense_quanquangu}, the maximum perturbations allowed are $\epsilon=0.3$ on MNIST and Fashion-MNIST, and $\epsilon=\frac{16}{255}$ on CIFAR-10 and CIFAR-100 for both threat and defense models.

\begin{figure}[!t]
\centering
\subfigure[Visualization]{\label{fig_visual_bvd}\includegraphics[width=4.2cm]{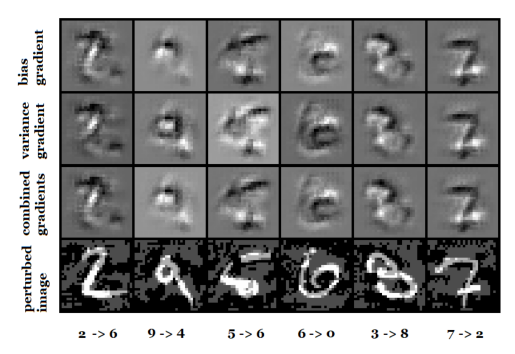}}
\subfigure[Bias-variance curve]{\label{curve_bvd_cnn}\includegraphics[width=4cm]{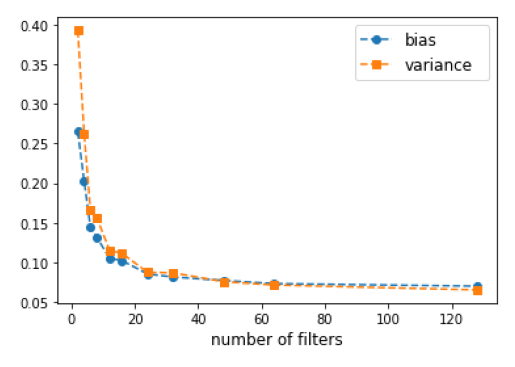}}
\vspace{-5mm}
\caption{Bias-variance analysis on MNIST with (a) visualizations of bias, variance, bias+variance, and perturbed images, and (b) Bias-variance curve w.r.t. the CNN model complexity}\label{fig:fig_mnist}
\vspace{-2mm}
\end{figure}

\begin{table}[!t]
\centering
\small
\scalebox{0.83}{
\setlength\tabcolsep{1.5pt}
\begin{tabular}{|l|ccc|ccc|}
\toprule
\multirow{2}{*}{Method}       & \multicolumn{3}{c|}{IID}          & \multicolumn{3}{c|}{non-IID}   \\ \cmidrule{2-7}
                        & Clean  & FGSM & PGD-20 & Clean & FGSM & PGD-20 \\ \midrule
Centralized         & \bf 0.991$_{\pm 0.000}$ & 0.689$_{\pm 0.000}$ & 0.182$_{\pm 0.000}$  & n/a & n/a & n/a \\
FedAvg        & 0.989$_{\pm \text{0.001}}$ & 0.669$_{\pm 0.009}$ & 0.267$_{\pm 0.014}$   & \bf 0.980$_{\pm 0.002}$ & 0.491$_{\pm 0.067}$ & 0.158$_{\pm 0.074}$ \\
FedAvg\_AT    & 0.988$_{\pm 0.000}$ & 0.802$_{\pm 0.001}$ & 0.512$_{\pm 0.042}$   & 0.974$_{\pm 0.005}$ & 0.649$_{\pm 0.066}$ & 0.363$_{\pm 0.066}$ \\
Fed\_Bias     & 0.986$_{\pm 0.000}$ & 0.812$_{\pm 0.009}$ & 0.583$_{\pm 0.036}$   & 0.971$_{\pm 0.004}$ & 0.679$_{\pm 0.040}$ & 0.394$_{\pm 0.103}$ \\
Fed\_Variance & 0.985$_{\pm 0.001}$ & 0.803$_{\pm 0.007}$ & 0.572$_{\pm 0.019}$   & 0.973$_{\pm 0.005}$ & 0.684$_{\pm 0.004}$ & 0.395$_{\pm 0.049}$ \\
Fed\_BVA      & 0.986$_{\pm 0.001}$ & 0.818$_{\pm 0.003}$ & 0.613$_{\pm 0.020}$   & 0.969$_{\pm 0.002}$ & 0.705$_{\pm 0.009}$ & 0.469$_{\pm 0.031}$ \\ \midrule
EAT           & 0.981$_{\pm 0.000}$ & \bf0.902$_{\pm 0.001}$ & 0.811$_{\pm 0.004}$   & 0.972$_{\pm 0.002}$ & 0.789$_{\pm 0.016}$ & 0.415$_{\pm 0.035}$ \\
EAT+Fed\_BVA  & 0.980$_{\pm 0.001}$ & 0.901$_{\pm 0.006}$ & \bf0.821$_{\pm 0.013}$   & 0.965$_{\pm 0.005}$ & \bf0.811$_{\pm 0.020}$ & \bf0.670$_{\pm 0.014}$ \\ \bottomrule
\end{tabular}}
\caption{Accuracy of MNIST under white-box attacks in IID and non-IID settings}
\label{tableResults:MNIST-IID-NONIID}
\vspace{-8mm}
\end{table}

\begin{table}[!t]
\centering
\small
\scalebox{0.83}{
\setlength\tabcolsep{1.5pt}
\begin{tabular}{|l|ccc|ccc|}
\toprule
\multirow{2}{*}{Method}       & \multicolumn{3}{c|}{IID}          & \multicolumn{3}{c|}{non-IID}   \\ \cmidrule{2-7}
& Clean  & FGSM   & PGD-20 &   Clean & FGSM & PGD-20 \\ \midrule
Centralized         & \bf0.882$_{\pm 0.000}$ & 0.229$_{\pm 0.000}$ & 0.009$_{\pm 0.000}$  & n/a & n/a & n/a \\
FedAvg           & 0.877$_{\pm 0.001}$ & 0.300$_{\pm 0.021}$  & 0.036$_{\pm 0.016}$  & {\bf0.804}$_{\pm 0.013}$ & 0.193$_{\pm 0.036}$  & 0.017$_{\pm 0.003}$ \\
FedAvg\_AT & 0.866$_{\pm 0.001}$ & 0.490$_{\pm 0.021}$ &    0.139$_{\pm 0.011}$ &  0.730$_{\pm 0.023}$ & {0.445}$_{\pm 0.065}$ & 0.087$_{\pm 0.042}$ \\
Fed\_Bias     & 0.862$_{\pm 0.001}$ & 0.505$_{\pm 0.015}$ &   0.159$_{\pm 0.003}$ &   0.709$_{\pm 0.025}$  & 0.460$_{\pm 0.038}$ &   {0.115}$_{\pm 0.054}$ \\
Fed\_Variance & 0.862$_{\pm 0.002}$ & 0.496$_{\pm 0.012}$ &   0.157$_{\pm 0.017}$ &   0.719$_{\pm 0.036}$ & 0.499$_{\pm 0.081}$ &   \bf0.120$_{\pm 0.038}$ \\
Fed\_BVA      & 0.862$_{\pm 0.003}$ & {0.528}$_{\pm 0.016}$   &   {0.180}$_{\pm 0.027}$   & 0.710$_{\pm 0.045}$ & 0.495$_{\pm 0.030}$ &   0.093$_{\pm 0.028}$ \\ \midrule
EAT              & 0.860$_{\pm 0.005}$ & \bf0.773$_{\pm 0.029}$ &   0.103$_{\pm 0.013}$   & 0.791$_{\pm 0.012}$ & 0.597$_{\pm 0.033}$ &  0.027$_{\pm 0.023}$ \\
EAT+Fed\_BVA  & 0.838$_{\pm 0.009}$ & 0.715$_{\pm 0.011}$ &   \bf0.226$_{\pm 0.006}$   &0.735$_{\pm 0.020}$ &\bf 0.632$_{\pm 0.015}$ &   0.106$_{\pm 0.039}$ \\ \bottomrule
\end{tabular}}
\caption{Accuracy of Fashion-MNIST under white-box attacks in IID and non-IID settings}
\label{tableResults:FashionMNIST-IID-NONIID}
\vspace{-8mm}
\end{table}

\begin{table}[!t]
\centering
\small
\scalebox{0.83}{
\setlength\tabcolsep{1.5pt}
\begin{tabular}{|l|ccc|ccc|}
\toprule
\multirow{2}{*}{Method}       & \multicolumn{3}{c|}{CIFAR-10}          & \multicolumn{3}{c|}{CIFAR-100}   \\ \cmidrule{2-7}
& Clean  & FGSM   & PGD-20 &   Clean & FGSM & PGD-20 \\ \midrule
Centralized       &{\bf 0.903}$_{\pm 0.003}$ &0.288$_{\pm 0.001}$ & 0.074$_{\pm 0.005}$     & {\bf 0.741}$_{\pm 0.003}$ & 0.166$_{\pm 0.012}$   & 0.032$_{\pm 0.003}$ \\
FedAvg       &0.890$_{\pm 0.002}$ &0.225$_{\pm 0.022}$  &0.062$_{\pm 0.008}$ &     {0.730}$_{\pm 0.003}$ & 0.161$_{\pm 0.009}$   & 0.035$_{\pm 0.006}$ \\
FedAvg\_AT   &0.890$_{\pm 0.003}$ &0.280$_{\pm 0.021}$ & 0.099$_{\pm 0.014}$ &     0.707$_{\pm 0.003}$ & 0.162$_{\pm 0.006}$ &  0.048$_{\pm 0.003}$ \\
Fed\_Bias    &0.890$_{\pm 0.004}$ &0.280$_{\pm 0.018}$  &0.103$_{\pm 0.012}$    & 0.702$_{\pm 0.002}$ & 0.163$_{\pm 0.005}$ &   0.061$_{\pm 0.003}$ \\
Fed\_Variance&0.889$_{\pm 0.001}$ &0.267$_{\pm 0.014}$  &0.092$_{\pm 0.009}$     & 0.710$_{\pm 0.007}$ & 0.161$_{\pm 0.005}$ &   0.045$_{\pm 0.016}$ \\
Fed\_BVA     &{0.889}$_{\pm 0.003}$ &{0.286}$_{\pm 0.013}$  &{0.104}$_{\pm 0.012}$     & 0.709$_{\pm 0.003}$ & {0.163}$_{\pm 0.007}$ &   {0.062}$_{\pm 0.005}$ \\ \midrule
EAT          &0.833$_{\pm 0.003}$ &0.596$_{\pm 0.003}$  &0.561$_{\pm 0.002}$ &     0.661$_{\pm 0.001}$ & 0.267$_{\pm 0.002}$ &   0.188$_{\pm 0.001}$ \\
EAT+Fed\_BVA &0.833$_{\pm 0.003}$ &\bf 0.598$_{\pm 0.002}$  &\bf 0.564$_{\pm 0.003}$     & 0.657$_{\pm 0.002}$ & \bf 0.272$_{\pm 0.003}$  &\bf 0.211$_{\pm 0.002}$ \\ \bottomrule
\end{tabular}}
\caption{Accuracy of CIFAR-10 and CIFAR-100 under white-box attacks}
\label{tableResults:CIFAR}
\vspace{-6mm}
\end{table}

\vspace{-2mm}
\subsection{Main Results}\label{sec:main_results}
To analyze the properties of our proposed {\tt \model} framework, we present two visualization plots on MNIST using a trained CNN model where the bias and variance are both calculated on the training examples. In Fig.~\ref{fig_visual_bvd}, we visualize the extracted gradients using adversarial attack from bias, variance, and bias-variance. Notice that the gradients of bias and variance are similar but with subtle differences in local pixel areas. However, according to Theorem~\ref{thm:empirical_estimation}, the gradient calculation of these two are quite different: bias requires the target label as input, but variance only needs the model output and main prediction. From another perspective, we also investigate the bias-variance magnitude relationship with varying model complexity. As shown in Fig.~\ref{curve_bvd_cnn}, with increasing model complexity (more convolutional filters in CNN), both bias and variance decrease. This result is different from the double-descent curve or bell-shape variance curve claimed in~\cite{belkin2019reconciling, yang2020rethinking}. The reasons are twofold: First, their bias-variance definitions are from the MSE regression decomposition perspective, whereas our decomposition utilizes the concept of the main prediction, and the generalization error is decomposed from the classification perspective; Second, their implementations only evaluate the bias and variance using training batches on one central model and thus is different from the definition which requires the variance to be estimated from multiple sub-models (in our scenario, client models).

\begin{figure}[!t]
\centering
\vspace{-5mm}
\subfigure[Convergence]{\label{fig_convergence_fashion}\includegraphics[width=2.9cm]{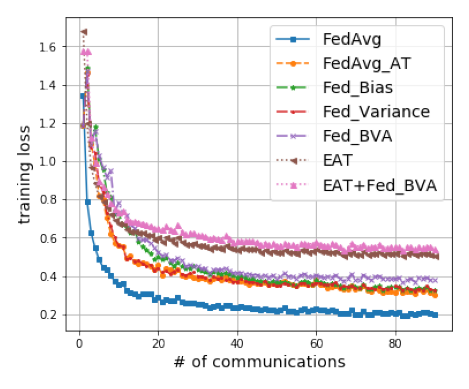}}
\subfigure[Performance]{\label{fig_performance_fashion}\includegraphics[width=2.9cm]{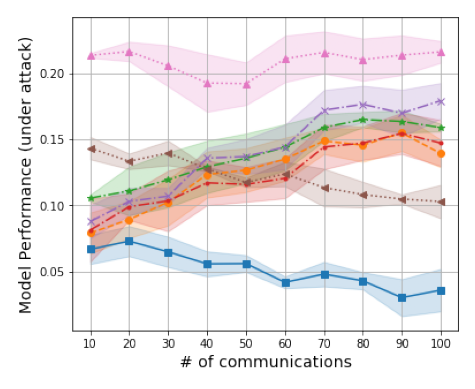}}
\subfigure[Efficiency]{\label{fig_efficiency_fashion}\includegraphics[width=2.5cm]{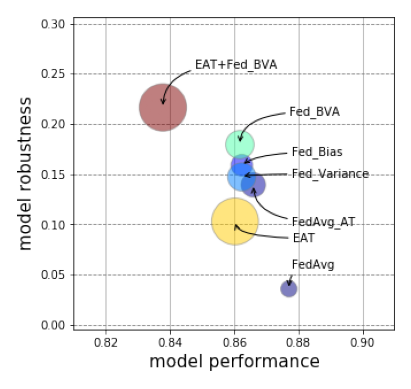}}
\vspace{-5mm}
\caption{Model analysis on Fashion-MNIST (PGD-20 attack)}\label{fig:fig_fashion}
\vspace{-2mm}
\end{figure}

The convergence plot of all baselines is presented in Fig.~\ref{fig_convergence_fashion}. We observe that {\tt FedAvg} has the best convergence, and all robust training will have a slightly higher loss upon convergence. This matches the observations in~\citep{madry2018towards} which states that training performance may be sacrificed in order to provide robustness for small capacity networks. For the model performance shown in Fig.~\ref{fig_performance_fashion}, we observe that the aggregation of federated learning is vulnerable to adversarial attacks since both {\tt FedAvg} and {\tt EAT} have decreased performance with an increasing number of server-client communications. Other baselines that utilized asymmetrical communications have increasing robustness with more communication rounds although only a small number of perturbed examples ($n_s=64$) are transmitted. We also observe that when communication rounds reach 40, {\tt \model} starts to outperform {\tt EAT} while the latter is even more resource-demanding than {\tt \model} (shown in Fig.~\ref{fig_efficiency_fashion}, where the pie plot size represents the running time). Overall, bias-variance based adversarial training via asymmetric communication is both effective and efficient for robust federated learning.

For the comprehensive experiments in Table~\ref{tableResults:MNIST-IID-NONIID} and  Table~\ref{tableResults:FashionMNIST-IID-NONIID}, it is easy to verify that our proposed model outperforms all other baselines regardless of the source of the perturbed examples (i.e., locally generated like {\tt EAT+\model} or asymmetrically transmitted from the server like {\tt \model}). In this case, BV-FGSM (see Eq. (\ref{BVD-FGSM})) is used to generate the adversarial examples $\hat{\mathcal{D}}_s$ during model training for robust federated learning. Compared to standard robust federated learning {\tt FedAvg\_AT}, the performance of {\tt \model} against adversarial attacks still increases $4\% - 13\%$ and $2\% - 9\%$ on IID and non-IID settings respectively, although {\tt \model} is theoretically suitable for the cases that clients have IID samples. In Table~\ref{tableResults:CIFAR}, we observe a similar trend where {\tt \model} outperforms {\tt FedAvg\_AT} on CIFAR-10 and CIFAR-100 (with $0.2\% - 10\%$ increases) when defending different types of adversarial examples. Compared to the strong local adversarial training baseline {\tt EAT}, we also observe a maximum $13\%$ accuracy increase when applying its bias-variance oriented baseline {\tt EAT+\model}. Overall, the takeaway is that without local adversarial training, using a bias-variance based robust learning framework could outperform other baselines for defending FGSM and PGD attacks. When local adversarial training is allowed (e.g., the client device has powerful computation ability), using bias-variance robust learning with local adversarial training will mostly have the best robustness.

\vspace{-2mm}
\subsection{Ablation Studies}
% We also conducted various additional studies, including: (1) comparison of efficiency and effectiveness of {\tt \model} using CE loss and MSE loss; (2) comparison of single-step {\tt \model} and multi-step {\tt \model} in terms of the generation of $\hat{\mathcal{D}}_s$; (3) three training scenarios of {\tt \model} that use client-specific adversarial examples or universal adversarial examples.
% ; (4) black-box attacking transferability among various models on all four data sets under multiple settings.
% \vspace{-1mm}
\subsubsection{MSE loss v.s. CE loss}\label{sec:mse_ce}
Both cross-entropy (CE) and mean squared error (MSE) loss functions could be used for training a neural network model. In our paper, the loss function of neural networks determines the derivation of bias and variance terms used for producing the adversarial examples. We experimentally compare the CE and MSE loss functions in our framework. Table~\ref{mse_ce} reports the adversarial robustness of our federated framework w.r.t. FGSM attack ($\epsilon=0.3$) on MNIST with IID setting. It is observed that (1) our framework with MSE has a significantly larger running time; (2) the robustness of our framework with MSE becomes slightly weaker, which might be induced by the weakness of MSE in training neural networks under the classification setting.

\begin{table}[!t]
\centering
\small
\begin{tabular}{|l|c|ccc|}
\toprule
\multirow{2}{*}{Loss} & \multirow{2}{*}{Clean} & \multicolumn{3}{c|}{\model{}} \\ \cmidrule{3-5}
  &   & BiasOnly        & VarianceOnly     & BVA    \\ \midrule
CE     & 0.588$_{({\bf 38.13s})}$          & {\bf 0.763}$_{\bf (47.58s)}$ & {\bf 0.759}$_{\bf (63.46s)}$  & {\bf 0.776}$_{\bf (63.67s)}$  \\
MSE               & {\bf 0.601}$_{(39.67s)}$          & 0.711$_{(65.03s)}$ & 0.711$_{(162.40s)}$ & 0.712$_{(179.60s)}$ \\ \bottomrule
\end{tabular}
\caption{Accuracy of training with different loss functions and running time (second/epoch)}\label{mse_ce}
\vspace{-8mm}
\end{table}

\vspace{-1mm}
\subsubsection{BV-PGD v.s. BV-FGSM}
Our bias-variance attack could be naturally generalized to any gradient-based adversarial attack algorithms when the gradients of bias $B(\cdot)$ and variance $V(\cdot)$ w.r.t. $x$ are tractable to be estimated from clients' models learned on finite training sets. Here, we empirically compare the adversarial robustness of the proposed framework trained with bias-variance guided PGD (BV-PGD) adversarial examples and bias-variance guided FGSM (BV-FGSM) adversarial examples. Table~\ref{fgsm_pgd} provides our results on w.r.t. FGSM and PGD attacks ($\epsilon=0.3$) on MNIST with IID and non-IID settings. Compared to FedAvg, our framework \model{} with either BV-FGSM or BV-PGD could largely improve the model robustness against adversarial noise. Furthermore, BV-PGD could potentially improve white-box robustness on multi-step attacks, but it is often more computationally demanding. As a comparison, BV-FGSM is more robust against single-step attacks.

\begin{table}[!t]
\centering
\small
\setlength\tabcolsep{2.7pt}
\begin{tabular}{|l|ccc|ccc|}
\toprule
\multirow{2}{*}{Method}   & \multicolumn{3}{c|}{IID}          & \multicolumn{3}{c|}{non-IID}      \\ \cmidrule{2-7}
  & FGSM   & PGD-10 & PGD-20 &  FGSM   & PGD-10 & PGD-20 \\ \midrule
FedAvg                 & 0.588 & 0.620 & 0.205 &  0.147 & 0.525 & 0.089 \\ 
\model{}$_{\text{(BV-FGSM)}}$     & {\bf 0.776} & 0.793 & 0.570  & {\bf 0.670} & 0.695 & 0.472 \\
\model{}$_{\text{(BV-PGD)}}$      & 0.757&    {\bf 0.840}    &   {\bf 0.632}   & 0.659 & {\bf 0.784} & {\bf 0.575} \\ \bottomrule
\end{tabular}
\caption{Comparison of training with BV-PGD and BV-FGSM}\label{fgsm_pgd}
\vspace{-8mm}
\end{table}

\begin{table}[!t]
\centering
\small
\begin{tabular}{|l|ccc|ccc|}
\toprule
\multirow{2}{*}{Strategy}   & \multicolumn{3}{c|}{IID}          & \multicolumn{3}{c|}{non-IID}      \\ \cmidrule{2-7}
& FGSM            & PGD-10          & PGD-20          &  FGSM            & PGD-10         & PGD-20          \\ \midrule
\textbf{S1} & \textbf{0.745} & \textbf{0.743} & \textbf{0.450} &  \textbf{0.529} & 0.677          & \textbf{0.433} \\
\textbf{S2} & 0.743          & 0.731          & 0.436        &  0.513         & \textbf{0.680} & 0.432          \\
\textbf{S3} & 0.730          & 0.704          & 0.400          &  0.495          & 0.657          & 0.380    \\ \bottomrule     
\end{tabular}
\caption{Robust training with different strategies on MNIST}\label{differ_scenarios}
\vspace{-5mm}
\end{table}

\vspace{-1mm}
\subsubsection{Alternative Training Strategies for \model}
Our {\tt \model{}} framework maximizes the overall generalization error induced by bias and variance from different clients for the adversarial examples generation. Under this setting, the generated adversarial examples on the server are shared with all the clients for local adversarial training. In particular, we found that when using the CE loss, the estimated gradients of bias and variance can be considered as the average of clients' local gradients over input $x$ (see Subsection~\ref{BVD_gradients}). This motivates us to consider several alternative training strategies by generating client-specific adversarial examples on the server. 
% To be more specific, we have the following three training strategies:
\begin{itemize}[leftmargin=*,noitemsep,nolistsep]
    \item {\bf S1:} We generate the adversarial examples $\hat{\mathcal{D}}_s$ to maximize the bias and variance from all clients' predictions. In this case, the generated adversarial examples on the server will be shared with the local clients. This is the strategy used in our {\tt \model{}} algorithm. It guarantees the minimization of generalization error from the perspective of bias-variance decomposition, thus leading to an adversarially robust federated learning model.
    \item {\bf S2:} The bias-variance decomposition with CE loss indicates that we generate the client-specific adversarial examples as
    \begin{equation*}
    \begin{aligned}
        \nabla_x B_k(x; w_k) &= \nabla_x L(f_{\mathcal{D}_k}(x;w_k), t) \\
        \nabla_x V_k(x; w_k) &= \sum_{j=1}^C (\log{y_m^{(j)}} + 1) \nabla_{x} f_{\mathcal{D}_k}^{(j)}(x;w_k)
    \end{aligned}
    \end{equation*}
    where $x\in \mathcal{D}_s$ and $k\in \{1,...,K\}$. If we using FGSM for attacking, the adversarial example on the $k^{\text{th}}$ client is:
    \begin{equation*}
        \hat{x}_{\mathrm{BV-FGSM}}^k := x + \epsilon \cdot \mathrm{sign}\left( \nabla_x \left(B_k(x; w_k) + \lambda V_k(x; w_k) \right) \right)
    \end{equation*}
    Note that the gradient estimate of client-specific variance also relies on the main prediction $y_m$. But in this case, it allows every client to have different adversarial examples $\hat{\mathcal{D}}_s$. Intuitively, this training strategy further decomposes the bias and variance into individual client-specific terms.
    \item {\bf S3:} Another training strategy is to use every local client model to generate the adversarial examples on the server individually as follows.
    \begin{equation*}
        \nabla_x B_k(x; w_k) = \nabla_x L(f_{\mathcal{D}_k}(x;w_k), t) \quad  \forall x\in \mathcal{D}_s \quad k\in \{1,...,K\}
    \end{equation*}
    Similarly, we can have: 
    \begin{equation*}
        \hat{x}_{\mathrm{BV-FGSM}}^k := x + \epsilon \cdot \mathrm{sign}\left( \nabla_x B_k(x; w_k) \right)
    \end{equation*}
    It is a special case of {\bf S2} with $\lambda=0$. In this case, every client will only use its own model parameters to generate the client-specific adversarial examples on the server. 
\end{itemize}
We conduct the ablation study to compare different training strategies in our {\tt \model{}} framework. In this case, we use $K=10$ clients with fraction $F=1$ and local epoch $E=5$. Other hyper-parameters and model architecture settings are the same as in our previous experiments. Table~\ref{differ_scenarios} provides the performance of adversarial robustness using our {\tt \model{}} framework on MNIST with both IID and non-IID settings. It is observed that {\tt \model{}} with {\bf S1} has the best robustness in most cases compared to other heuristic training strategies {\bf S2} and {\bf S3}. This indicates that bias and variance provide better direction to generate the adversarial examples for robust training. In contrast, detecting the adversarial examples with individual directions in {\bf S2} for each client might be sub-optimal.

\begin{table}[!t]
\centering
\small
\scalebox{0.85}{
\setlength\tabcolsep{4pt}
\begin{tabular}{|l|cccc|cccc|}
\toprule
\bf CIFAR-10                & \multicolumn{4}{c|}{\textbf{Source} (FGSM attack)} & \multicolumn{4}{c|}{\textbf{Source} (PGD-20 attack)}  \\ \cmidrule{1-9}
\textbf{Target}   & R & V & X & M & R & V & X & M \\\midrule
FedAvg            &0.707 &0.688 &0.689 &0.793  &0.611 &0.623 &0.597 &0.787\\
FedAvg\_AT        &0.742 &{0.710} &0.720 &0.808  &{0.695} &{0.670} &0.661 &0.808\\
Fed\_Bias         &0.740 &0.703 &0.715 &0.799  &0.690 &0.667 &0.654 &0.799\\
Fed\_Variance     &0.738 &0.704 &0.719 &{0.810}  &0.677 &0.656 &0.648 &0.809\\
Fed\_BVA          &{0.744} &0.706 &{0.722} &0.809  &0.693 &0.669 &{0.664} &{0.809}\\ \midrule
EAT               &0.821 &0.806 &0.815 &0.823  &0.819 &0.808 &\bf0.813 &0.822\\
EAT+Fed\_BVA      &\bf0.828 &\bf0.808 &\bf 0.817 &\bf0.828  &\bf0.825 &\bf0.809 &0.812 &\bf0.829  \\ \bottomrule
\end{tabular}}
\caption{CIFAR-10 accuracy towards black-box adversarial examples on transferability between models (R: ResNet18; V: VGG11; X: Xception; M: MobileNetV2)}
\label{bbresults:cifar_10}
\vspace{-6mm}
\end{table}

\vspace{-4mm}
\subsection{Against Other Attacks}
\vspace{-1mm}
\subsubsection{Black-box Attacks}
Using black-box attack, we test the transferability of single-step attack (i.e., FGSM~\cite{goodfellow2014explaining}) and multi-step attack (i.e., PGD~\cite{kurakin2016adversarial}) on various federated learning baseline models. For CIFAR-10 data set, we use ResNet18~\citep{resnet_he_2016}, VGG11~\citep{vgg_zisserman_2015}, Xception~\citep{xception_chollet_2017}, and MobileNetV2~\citep{mobilenetv2_sandler_2018} as the source threat models for generating the single-step and multi-step adversarial examples. The black-box transfer attacking results are shown in Table~\ref{bbresults:cifar_10}. We observe that without any adversarial training, {\tt FedAvg} will suffer a maximum of $28\%$ accuracy lost. For comparison, the robust federated learning model with global transmitted perturb samples (i.e., {\tt FedAvg\_AT}, {\tt Fed\_Bias}, {\tt Fed\_Variance}, {\tt Fed\_BVA}) will have increase robustness with a maximum of $23\%$ accuracy drop on best baselines. For the computation-demanding local robust training methods (i.e., {\tt EAT} and {\tt EAT+Fed\_BVA}),  the maximum accuracy drop is only $3\%$, respectively. Thus, it is straightforward to see that CIFAR-10 is more vulnerable to multi-step black-box adversarial attacks, but the local adversarial training methods could improve its robustness.

\begin{figure}[!t]
\centering
\subfigure[FGSM \& Sign-flipping]{\label{fgsm_sign}\includegraphics[width=3.5cm]{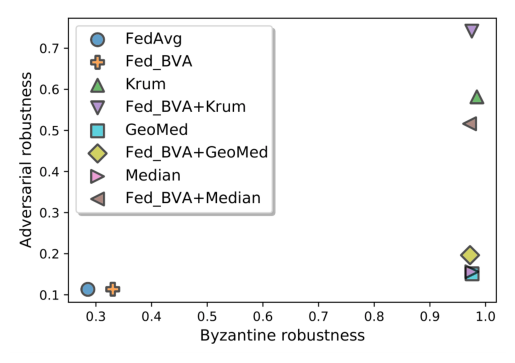}}
\subfigure[PGD-20 \& Additive noise]{\label{pgd_noise}\includegraphics[width=3.5cm]{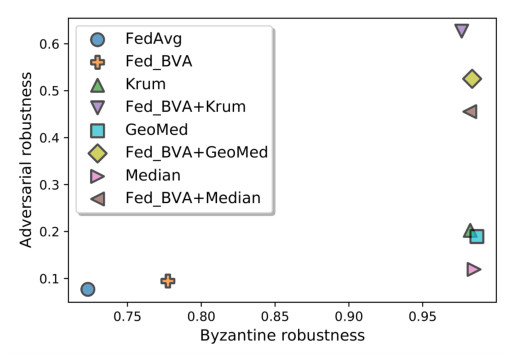}}
\vspace{-5mm}
\caption{Adversarial and Byzantine robustness on MNIST with (a) FGSM attack on the test set and sign-flipping attack on local model updates, and (b) PGD-20 attack on the test set and additive noise attack on local model updates}\label{fig:byzantine}
\vspace{-2mm}
\end{figure}

\vspace{-3mm}
\subsubsection{Byzantine Attacks}\label{sec:byzantine_attacks}
The proposed {\tt Fed\_BVA} is flexible to incorporate with Byzantine-robust aggregation variants, in order to improve both the adversarial robustness against the corrupted test data set and the Byzantine robustness against the corrupted local model updates. Here we use sign-flipping attack~\cite{li2020learning} and additive noise attack~\cite{li2020learning} to manipulate the local model updates of 40\% clients. Then we adopt the Byzantine-robust aggregation mechanisms in our Algorithm~\ref{algorithm:RobustFed}. In this case, we use three popular mechanisms, i.e., Krum~\cite{blanchard2017machine}, Median~\cite{yin2018byzantine} and GeoMed~\cite{chen2017distributed}, to aggregate the local model updates. We report the results in Figure~\ref{fig:byzantine}. It is observed that {\tt Fed\_BVA} is robust against adversarial and Byzantine attacks, when using Krum as our aggregation strategy.

\vspace{-2mm}
\subsection{Parameter Analysis}\label{sec:Parameter_Analysis}
% We perform parameter analysis regarding the key hyper-parameters.

\vspace{-1mm}
\subsubsection{Number of shared perturbed samples $n_s$}\label{sec:shared_samples}
\begin{figure}[!t]
\centering
\subfigure[Clean training]{\label{fig:numShare_a}\includegraphics[width=2.8cm]{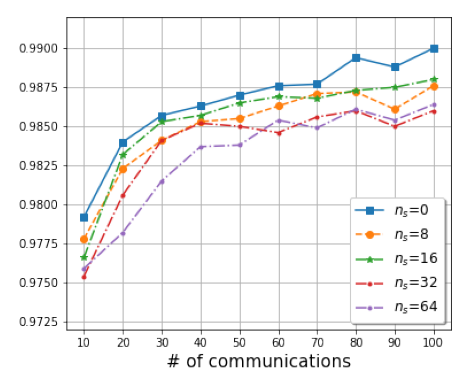}}
\subfigure[Under FGSM attack]{\label{fig:numShare_b}\includegraphics[width=2.7cm]{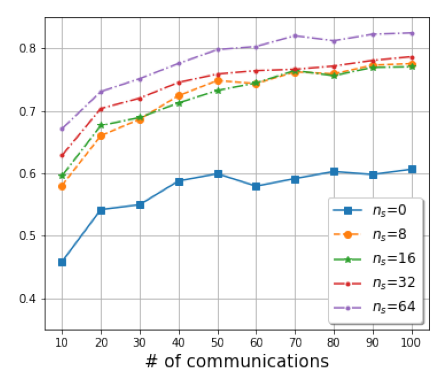}}
\subfigure[Under PGD-20 attack]{\label{fig:numShare_c}\includegraphics[width=2.7cm]{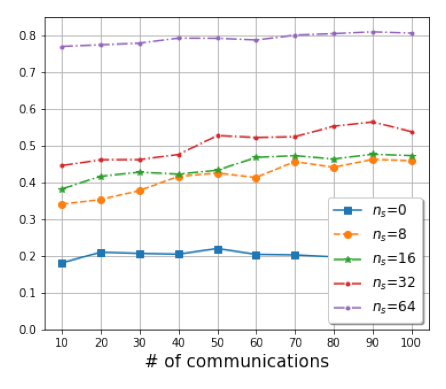}}
\vspace{-6mm}
\caption{Accuracy of \model{} w.r.t. number of shared data $n_s$}\label{fig:numShare}
\vspace{-6mm}
\end{figure}

From Fig.~\ref{fig:numShare}, we see that the accuracy of {\tt FedAvg} (i.e., $n_s=0$) has the best accuracy as we expected. For {\tt \model} with varying size of asymmetrical transmitted perturbed samples (i.e., $n_s=8, 16, 32, 64$), its performance drops slightly with increasing $n_s$ (on average drop of $0.05\%$ per plot). As a comparison, 
the robustness on the test set increases dramatically with increasing $n_s$ (the improvement ranges from $18\%$ to $22\%$ under FGSM attack and ranges from $15\%$ to $60\%$ under PGD-20 attack). However, choosing large $n_s$ would have high model robustness but also suffer from the high communication cost. In our experiments, we choose $n_s=64$ for MNIST for the ideal trade-off point. 

\vspace{-2mm}
\subsubsection{Momentum}
\begin{figure}[!t]
\centering
\subfigure[Clean training]{\label{fig:momentum_a}\includegraphics[width=2.8cm]{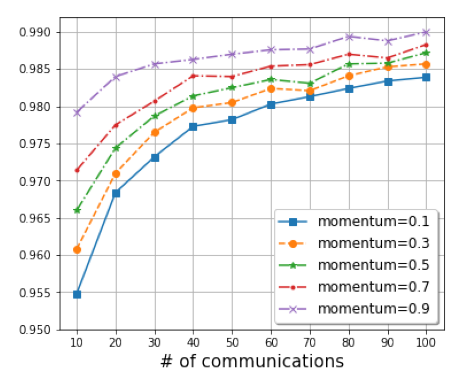}}
\subfigure[Under FGSM attack]{\label{fig:momentum_b}\includegraphics[width=2.7cm]{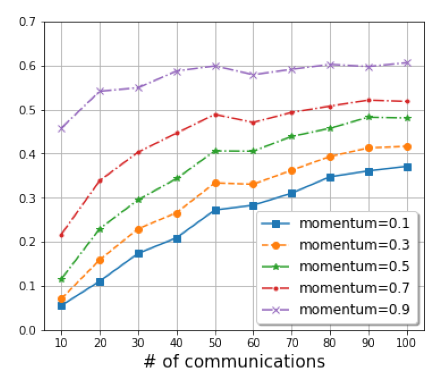}}
\subfigure[Under PGD-20 attack]{\label{fig:momentum_c}\includegraphics[width=2.7cm]{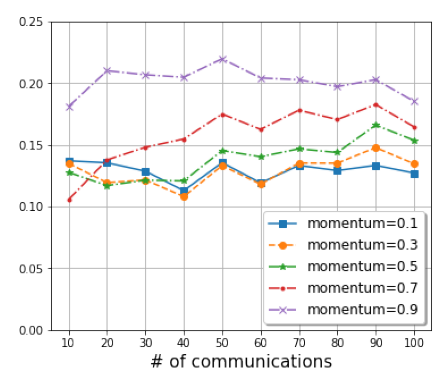}}
\vspace{-6mm}
\caption{Accuracy of \model{} w.r.t. momentum}\label{fig:momentum}
\vspace{-6mm}
\end{figure}

We also care about the choice of options in the SGD optimizer. Fig.~\ref{fig:momentum_a} shows that the accuracy of clean training monotonically increases when the momentum is varying from $0.1$ to $0.9$. Interestingly, we observe from Fig.~\ref{fig:momentum_b} and Fig.~\ref{fig:momentum_c} that the federated learning model is less vulnerable when momentum is large no matter whether the adversarial attack is FGSM or PGD-20 on $\mathcal{D}_{test}$. So we choose the momentum as $0.9$ for our experiments.

\vspace{-2mm}
\subsubsection{Local epochs $E$}
\begin{figure}[!t]
\centering
\subfigure[Clean training]{\label{fig:local_epoch_a}\includegraphics[width=2.7cm]{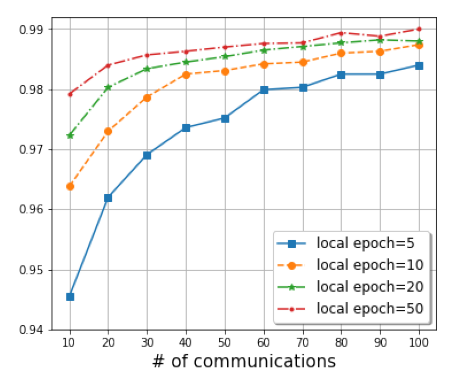}}
\subfigure[Under FGSM attack]{\label{fig:local_epoch_b}\includegraphics[width=2.7cm]{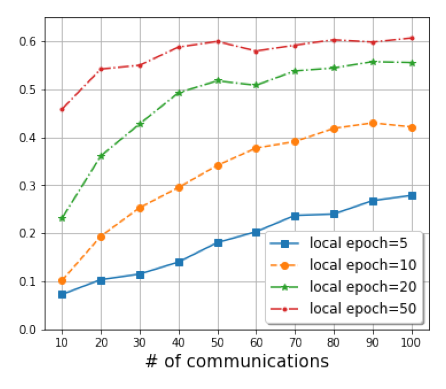}}
\subfigure[Under PGD-20 attack]{\label{fig:local_epoch_c}\includegraphics[width=2.7cm]{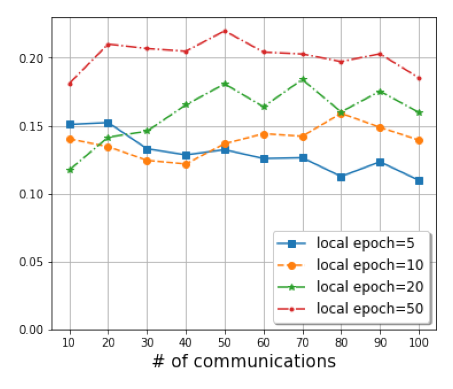}}
\vspace{-6mm}
\caption{Accuracy of \model{} w.r.t. local epoch}\label{fig:localepoch}
\vspace{-2mm}
\end{figure}

Another important factor of federated learning is the number of local epochs. In Fig.~\ref{fig:localepoch}, we see that more local epochs in each client lead to a more accurate aggregated server model in prediction. Its robustness against both FGSM and PGD-20 attacks on the test set $\mathcal{D}_{test}$ also has the best performance when the local epochs are large on the device. Hence, in our experiments, if the on-device computational cost is not very high (large data example size, deep models with many layers), we choose $E=50$. Otherwise, we will reduce $E$ to a smaller number accordingly.

\vspace{-2mm}
\section{Conclusion}\label{sec:conclusion}
In this paper, we proposed a novel robust federated learning framework, in which the loss incurred during the server's aggregation is dissected into a bias part and a variance part. Our approach improves the model robustness through adversarial training by supplying a few bias-variance perturbed samples to the clients via asymmetrical communications. Extensive experiments have been conducted where we evaluated its performance from various aspects on several benchmark data sets.

% We believe the further exploration of this direction will lead to more findings on the adversarial robustness of federated learning.
% \vspace{-2mm}
\begin{acks}
  This work is supported by National Science Foundation under Award No. IIS-1947203, IIS-2117902, IIS-2137468, and Agriculture and Food Research Initiative (AFRI) grant no. 2020-67021-32799/project accession no.1024178 from the USDA National Institute of Food and Agriculture. The views and conclusions are those of the authors and should not be interpreted as representing the official policies of the funding agencies or the government.
\end{acks}

\clearpage
\clearpage
\bibliographystyle{ACM-Reference-Format}
\balance
\bibliography{acmart}

\end{document}